\pdfoutput=1

\documentclass[11pt]{article}

\usepackage{acl}

\usepackage{times}
\usepackage{latexsym}
\usepackage{url}
\usepackage[T1]{fontenc}

\usepackage[utf8]{inputenc}

\usepackage{microtype}

\usepackage{inconsolata}
\usepackage{times}
\usepackage{booktabs}
\usepackage{todonotes}
\usepackage{latexsym}
\usepackage{float}
\usepackage{bbding}
\usepackage{pifont}
\usepackage{wasysym}
\usepackage{amssymb}
\usepackage{multirow}
\usepackage{makecell}
\usepackage{subcaption}

\usepackage{amsmath}
\usepackage{bm}
\usepackage{mathrsfs}
\usepackage{soul} 

\usepackage{tcolorbox}
\usepackage{xcolor}
\usepackage{enumitem}
\setitemize{itemsep=10pt,topsep=0pt,parsep=0pt,partopsep=0pt}

\usepackage{hyperref}

\title{An Examination of the Compositionality of Large Generative Vision-Language Models}

\author{
    \begin{tabular}{c}
    Teli Ma$^{\dagger, \diamond}$ \quad Rong Li$^{\dagger}$\quad Junwei Liang$^{\dagger , \ddag, \vartriangle}$ 
    \end{tabular}
    \\ %
    \begin{tabular}{c}
    $^\dagger$ AI Thrust, The Hong University of Science and Technology (Guangzhou) \\
    $^\ddag$ %
    Department of Computer Science and Engineering, \\The Hong Kong University of Science and Technology \quad \\
    $^\diamond$ Primary author \quad $^\vartriangle$ Corresponding author \\
    \end{tabular}
    \\ %
    \begin{tabular}{c}
    \texttt{\{tma184, rli335\}@connect.hkust-gz.edu.cn}  \\
    \texttt{junweiliang@hkust-gz.edu.cn} \\
    \end{tabular} 
    \vspace{2mm} \\ \\
}

\begin{document}

\maketitle

\begin{abstract}
With the success of Large Language Models (LLMs), many Generative Vision-Language Models (GVLMs) have been constructed via multimodal instruction tuning.
However, the performance of GVLMs in multimodal compositional reasoning remains under-explored. 
In this paper, we examine both the evaluation metrics ( VisualGPTScore, etc.) and current benchmarks for evaluating the compositionality of GVLMs. 
We identify the syntactical bias in current benchmarks, which is exploited by the linguistic capability of GVLMs. 
The bias renders VisualGPTScore an insufficient metric for assessing GVLMs.
To combat this, we first introduce a \textbf{SyntaxBias Score}, leveraging LLMs to quantify such bias for mitigation.
A challenging new task is subsequently added to evaluate the robustness of GVLMs against inherent inclination toward syntactical correctness.
Using the bias-mitigated datasets and the new task, we propose a novel benchmark, namely \textbf{S}ynt\textbf{A}ctically \textbf{DE}-biased benchmark (SADE).
Our study provides an unbiased benchmark for the compositionality of GVLMs, facilitating future research in this direction~\footnote{Code and dataset are available at \url{https://github.com/TeleeMa/SADE}.}.
\end{abstract}

\section{Introduction}
A surge of research on vision-language models (VLMs) has demonstrated success in a wide range of tasks, including zero-shot visual recognition~\cite{CLIP, clipadapter, detecting21k}, visual question answering~\cite{flamingo, pali}, and image-to-text retrieval~\cite{flamingo,openflamingo}. 
Previous Vision-Language Models (VLMs) have predominantly been developed using image-text contrastive (ITC) learning~\cite{CLIP,ALIGN,BLIP,blip2} and image-text matching (ITM)~\cite{tan2019lxmert, chen2020uniter, villa, ALBEF, zhang2021vinvl, kim2021vilt} frameworks, a category we term Encoder-based Vision-Language Models (EVLMs).
With the advent of large language models (LLMs) like ChatGPT, GPT-4~\cite{openai2023gpt4} and LLaMA~\cite{llama}, recent studies have extended the decoder-only architecture to multimodal settings, which is named Generative VLMs (GVLMs)~\cite{llava, minigpt4, otter, mplug, llama-adapter-v2, emu, instructblip}. The GVLMs deviate from the EVLMs in projecting visual features into the latent lexical space of LLMs, and leveraging the auto-regressive generative capacity to solve vision-language tasks. In the training process, most work follows the recipe of freezing the main body of visual encoders and LLMs, only updating the negligible parameters of projecting layers, which is also called ``bridge architecture"~\cite{rajesh2023bridging}.

Despite the emergence of research on GVLMs, the understanding of compositionality in GVLMs has remained an enigmatic black
box, with no thorough investigations conducted thus far. 
Previous research studies~\cite{winoground, vl-checklist, ARO, crepe, ray2023cola} in multimodal compositionality focus on establishing retrieval-based benchmarks for evaluating EVLMs on object relations and attribute understanding, order sensitiveness of sentence elements, and atom-level understanding. 
The EVLMs have demonstrated abilities to discriminate positive captions from negative ones based on the image-text similarity, where the disparities between the positive and negative captions are relatively subtle, such as ``an old person kisses a young person" and ``a young person kisses an old person"~\cite{winoground}. 

However, we observe there exists an underlying bias towards the LLM part of GVLMs in the evaluation of the aforementioned benchmarks. 
During the evaluation,
the log-likelihood-based scores are widely adopted to evaluate the generative models~\cite{gptscore, mmbench, visualgptscore, li2023reform} to estimate the conditional probabilities of specific generations. 
Following~\citet{visualgptscore}, we alias the log-likelihood score as VisualGPTScore.
We examine the current benchmarks for evaluating GVLMs with VisualGPTScore and find that:

\begin{itemize}
    \item Using VisualGPTScore to evaluate GVLMs is not sensitive to bags-of-words problems that broadly exist in the evaluation of EVLMs with similarity scores. The bags-of-words phenomenon during evaluation is due to the similarity-based metrics.
    
    \item VisualGPTScore sometimes prefers syntactical correctness rather than content-related correctness under the current benchmarks. It scores negative references with reasonable syntax but unrelated content higher than positive references. In contrast, EVLMs pay more attention to the correlation of visual content but are not sensitive to the order of tokens in references. 
    
    \item A prevalent syntactical bias is present in contemporary multimodal compositional reasoning benchmarks.%
    These benchmarks are tailored for assessing EVLMs, and the approach used to create negative references may not be effective for the evaluation of GVLMs.
\end{itemize}

    \noindent Based on these observations, our contributions include:
\begin{itemize}
    \item We quantitatively analyze the syntactical bias (namely SyntaxBias Score) that broadly exists in current benchmarks by leveraging LLMs.%
    \item With the SyntaxBias Score, we propose a SyntActically DE-biased benchmark (SADE) based on current benchmarks for a more robust multimodal compositionality evaluation. 
    We adopt multiple strategies to mitigate the syntactical bias in existing benchmarks.
    We also add a new challenging assessment in SADE to evaluate the content understanding across visual and language modalities. 
    \item The performance of several GVLMs is reported on SADE, as well as the robustness and faithfulness to human judgments.
\end{itemize}

\section{Background}

\subsection{Generative vision-language models}
In this paper, we define GVLMs as models that combine visual encoders with large language models (LLMs) trained on large text corpora.
The prevailing approach in recent research connects a frozen visual encoder with an LLM by training mapping layers on images-text pairs, followed by fine-tuning using multi-modal instructional data to facilitate multi-turn conversations~\cite{llava, llama-adapter-v2, minigpt4, instructblip, pandagpt, mmgpt, emu}. 
This approach is anchored in the idea of treating visual tokens the same as linguistic ones.
The visual tokens are mapped into a lexical embedding space and harnessed to generate textual content in an autoregressive manner.
Formally, given an image $I$ and the visual encoding $g(I)$ from encoders like Vision Transformer~\cite{vit}, the mapping process can be formulated as:
\begin{equation}
   \bm{z} = \mathbf{M} (g(I)), \bm{z}=\{z_1, z_2, ... , z_N\},
\end{equation}
where $N$ is the number of visual tokens and $\mathbf{M}$ is the mapping layers.
Different from EVLMs that utilize image-text contrastive (ITC) or image-text matching (ITM), the training objective of multi-modal autoregressive training is to maximize the log-likelihood of the next true token. 
Denote the tokenized instructions as $\bm{p}$ and the output words as $t_i, (1\leq i \leq K)$, the GVLM training objective is defined as:
\begin{equation}
    \mathop{{\rm max}}\limits_{\theta_M, \theta_{\sigma}} \sum\limits_{i=1}^{K} {\rm log} P(t_i|\bm{p}, \bm{z}, t_1, t_2, ... ,t_{i-1}; \theta_M, \theta_{\sigma})
\end{equation}
where $\theta_M$ refers to the learnable parameters of mapping layers $\mathbf{M}$ and $\theta_{\sigma}$ refers to other tunable parameters like adapter layers in LLaMA-Adapter V2~\cite{llama-adapter-v2}, or visual abstractor and LoRA in mPLUG-Owl~\cite{mplug}.

In comparison, the training objectives of EVLMs are based on the ITC or ITM loss between vision and language parts. Please refer to Appendix \ref{subsec:formulation} for formulations of EVLMs. 

\subsection{Vision-language compositionality}
Recent works on vision-language compositionality focus on introducing benchmarks to evaluate the EVLMs, mainly on CLIP~\cite{CLIP}. 
Winoground~\cite{winoground} is one of the pioneers in building benchmarks for multimodal compositionality, curating 400 test items to evaluate the pragmatics, symbolic and series factors of VLMs. 
Afterwards, several benchmarks have been proposed to challenge the objects, relations and attributes understanding of VLMs, including VL-CheckList~\cite{vl-checklist}, ARO~\cite{ARO}, CREPE~\cite{crepe}, VALSE~\cite{parcalabescu2021valse} and Cola~\cite{cola} \textit{etc.} 
These benchmarks are in the form of image-text retrieval, requiring the model to differentiate positive references from negative references based on the visual contents of the images.
See Fig~\ref{fig:example} in the Appendix for the details of the image-text retrieval format. 
SugarCrepe~\cite{hsieh2024sugarcrepe} is one of the most recent and similar works to ours. SugarCrepe utilizes Vera~\cite{liu2023vera} and TextAttack~\cite{morris2020textattack} to detect the plausibility and grammar gaps between positive and negative references. Then, it prompts ChatGPT to generate reasonable hard negative references to reduce bias. In comparison, we partially rely on the original benchmarks, focusing on the strategy of mitigating bias by filtering and modifying them based on our defined SyntaxBias Score. %
All the aforementioned benchmarks are curated for evaluating EVLMs, where similarity scores between images and references serve as the criteria for selecting references. Then, the accuracy of selecting positive samples across all data samples will be reported to assess the model's compositional understanding capability.

\subsection{Evaluation metrics for multimodal retrieval}
\label{sec:background}
Since previous benchmarks have been carefully curated for evaluating EVLMs, image-text similarity scores naturally emerge as the metric for assessing the compositional similarity between images and references. For generative models, an intuitive way is reference-based, measuring the quality of generated captions with metrics like BLEU~\cite{bleu}, METEOR~\cite{meteor}, ROUGE~\cite{rouge} and CIDEr~\cite{cider}. Among the reference-based metrics, BERTScore~\cite{bertscore} tackles superficial matching between captions and references in lexical expression, delving deeper into the semantic similarity matching.
GPTScore~\cite{gptscore} proposes to leverage emergent abilities of generative models to score generated texts. Inspired by GPTScore, recent works~\cite{visualgptscore, li2023reform, mmbench} measure the GVLMs using the log-likelihood of directly generating reference sentences conditioned on the image. We follow the \citet{visualgptscore} to abbreviate the kind of method as VisualGPTScore, which can be formulated as:
\begin{align}
\label{eq:visualgptscore}
    &{\rm VisualGPTScore}(\bm{r}|\mathcal{I}) \nonumber \\
    &= \sum\limits_{t=1}\limits^{m}w_t {\rm log} P(r_t|\bm{r}_{<t}, \bm{p}, \mathcal{I}; \theta_{GVLM})
\end{align}
where $\mathcal{I}$, $\bm{r}$, $\bm{p}$ represents the image, reference sentence and instructions. $\theta_{GVLM}$ refers to parameters of GVLMs and $w_t=\frac{1}{m}$. The VisualGPTScore is directly estimated conditioned on images and thus reference-free. In this work, we examine the VisualGPTScore and discuss the potential influence of using it in current benchmarks for vision-language compositionality.

\section{Experimental setup}
We introduce the configurations of experiments for the syntactical bias examination in this section. 

\subsection{Model choices} 
We leverage two state-of-the-art GVLMs, namely LLaVA~\cite{llava} and MiniGPT-4~\cite{minigpt4}, to conduct experiments. 
LLaVA is one of the first methods to project visual features into LLaMA~\cite{llama} latent space via multimodal instruction tuning. 
A linear projection layer and the parameters of the LLM are tuned on conversations, detailed descriptions, and complex reasoning datasets. 
MiniGPT-4~\cite{minigpt4} maps visual embeddings obtained from ViT and Q-Former~\cite{BLIP} into Vicuna~\cite{vicuna} via a linear projection layer. 
We adopt the model version of ``LLaVA-7B-v0" and ``Minigpt4-aligned-with-Vicuna7B" to evaluate.
However, we found that when using VisualGPTScore to evaluate compositionality, both models exhibited similar patterns. Therefore, for the sake of brevity, we only present the results for LLaVA.

\begin{figure}[htbp]
\centering
\includegraphics[width=1.0\linewidth,trim={0cm 0cm 0cm 0cm}]{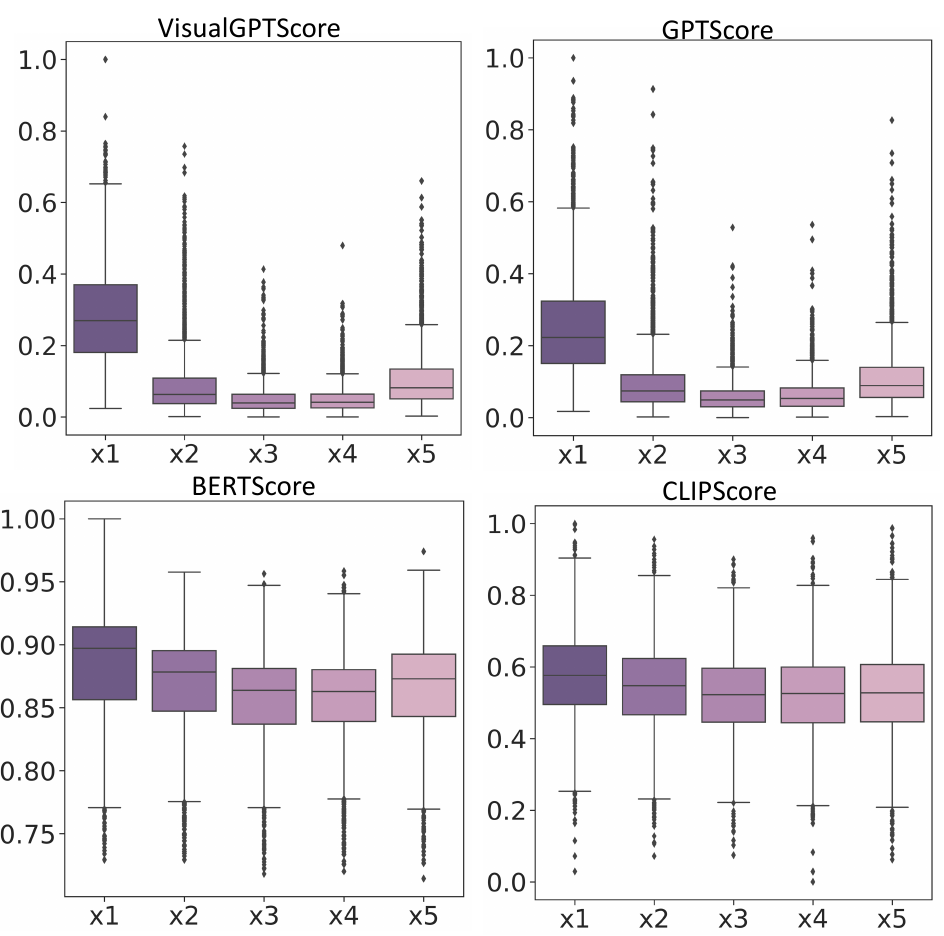}
\caption{Box plots of scaled score distributions for original (\texttt{x1}) and perturbed captions (\texttt{x2-x5,} \texttt{x2: shuffle nouns \& adj, x3: shuffle all but nouns \& adj, x4: shuffle within trigrams, x5: shuffle trigrams}). The distribution gap between the original captions and the shuffled captions is evident for the generative scores, 
while the contrastive score (BERTScore) is significantly less affected by the order of words. The CLIPScore sub-figure illustrates the distribution of similarity scores generated by the CLIP model, which is compared with the first three sub-figures of LLaVA-7B.
}
\vspace{-10pt}
\label{fig:bag_of_words}
\end{figure}

\subsection{Datasets}
We use Winoground~\cite{winoground}, VL-Checklist~\cite{vl-checklist}, ARO~\cite{ARO} and CREPE~\cite{crepe} in the evaluation analysis, totaling 52,189 images and 129,558 reference sentences. All benchmarks necessitate the model's selection of positive reference sentences from negative ones.
For Winoground, we report text score, image score and group score as the paper~\cite{winoground}. For other datasets, Recall@1 accuracy is reported.

\section{Evaluation Metric Examination}
\label{sec:evaluation metric}
\label{sec:evaluation}
VisualGPTScore measures the probability of generating specific references conditioned on the given images, as defined in Eqn.~\ref{eq:visualgptscore}. The generative evaluation method is based on the inherent attribute of GVLMs and used in image-text retrieval~\cite{visualgptscore, li2023reform, mmbench}. Since current benchmarks on VL compositions consists of image-text pairs, we follow \citet{visualgptscore} to utilize VisualGPTScore for evaluating the VL compositionality of GVLMs. 
In this section, our primary focus is to examine the bias of using VisualGPTScore in current benchmarks.

\subsection{Sensitivity to bags-of-words}

\begin{figure*}[h!]
\centering
\includegraphics[width=1.0\linewidth,trim={0cm 0cm 0cm 0cm}]{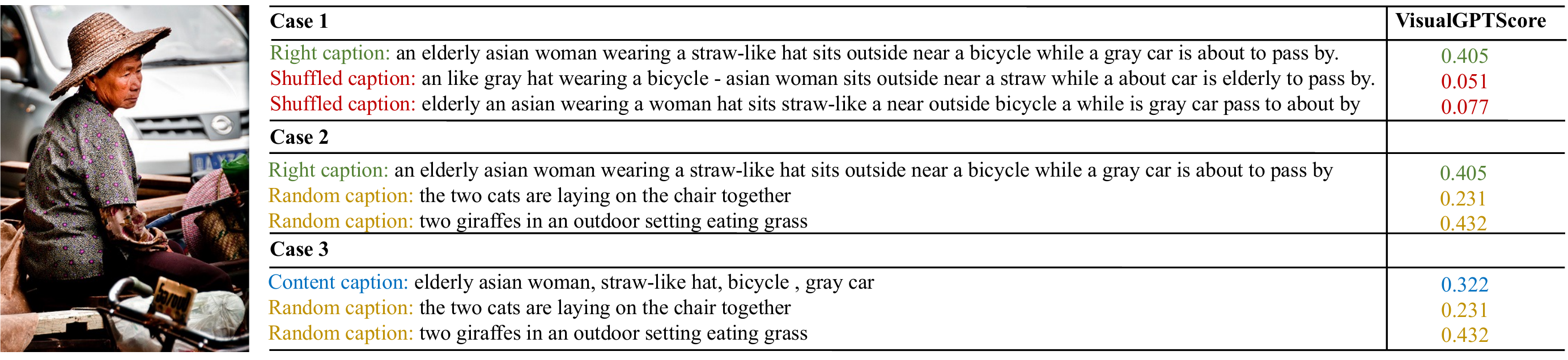}
\caption{
\textbf{An example of three \texttt{Cases} of captions we construct to validate the preference of syntax and contents.} \texttt{Right caption}: the original caption of the image, 
\texttt{Shuffled caption}: caption that the sentence elements are shuffled, \texttt{Random caption}: fluent and syntactically correct captions from other datasets (COCO), \texttt{Content caption}: caption that keeps only adjectives and nouns to keep the contents like objects and attributes. We present the normalized VisualGPTScore of every reference sentences in this example. The scores of the \texttt{Right caption} and \texttt{Content caption} may be lower compared to the \texttt{Random caption} (0.405, 0.322 vs. 0.432). This indicates that in this example, generative VLMs tend to prioritize syntactically correct sentences over ones that are more relevant to the content. }
\label{fig:syntax}
\end{figure*}

\begin{figure}[h!]
\centering
\includegraphics[width=0.8\linewidth,trim={0cm 0cm 0cm 0cm}]{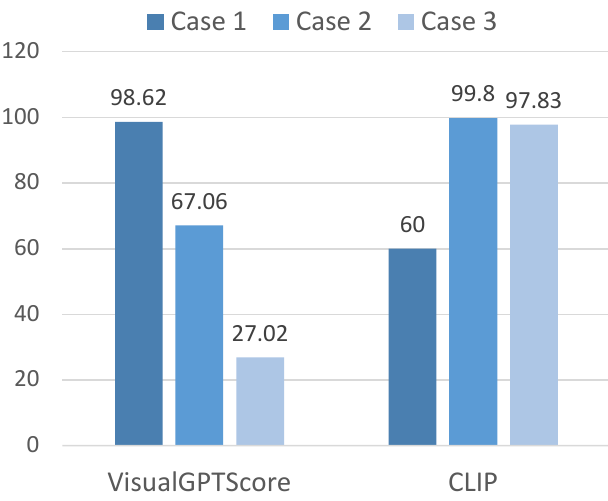}
\caption{
 We report the accuracy of VisualGPTScore based on LLaVA-7B and similarity score based on CLIP in the sampled 507 image-text pairs, each pair is consisted of three cases like the example in Fig.~\ref{fig:syntax}.
}
\label{fig:syntax_chart}
\end{figure}

Previous research works have pointed out that EVLMs suffer from the bags-of-words phenomenon when doing compositional reasoning due to the pre-training recipe of matching visual and textual data in instances-level~\cite{whenandwhy, whywinogroundhard}. 
However, we observe that the bags-of-words problem is not only related to the models, but also highly correlated to the evaluation metrics, and VisualGPTScore is not sensitive to the bags-of-words phenomenon.

We explore the influence of different metrics in sensitivity to the order of tokens in sentences for GVLMs. Following CREPE~\cite{crepe}, we randomly sample 2.5K image-text pairs from the COCO dataset~\cite{coco} and adopt the following strategies to shuffle the elements of captions: \textit{Shuffle only nouns \& adjectives}, \textit{Shuffle all but nouns \& adjectives}, \textit{Shuffle within trigrams}, \textit{Shuffle trigrams}. 
Then, we calculate the VisualGPTScore, GPTScore~\cite{gptscore} and BERTScore~\cite{bertscore} based on LLaVA-7B.
The distribution of normalized scores are shown in Fig.~\ref{fig:bag_of_words}, where \texttt{x1} represents positive references and \texttt{x2-x5} represents shuffled references,  respectively.

It can be observed that to the same model, LLaVA-7B, 
VisualGPTScore is similar to GPTScore, more sensitive to the order and structure of reference sentences compared with contrastive metric BERTScore. We also report the score distribution of the CLIP model using contrastive similarity (CLIPScore in Fig.~\ref{fig:bag_of_words}), which is similar to the distribution of BERTScore results on LLaVA-7B.
It implies the bags-of-words problem may be attributed to the evaluation metrics based on similarity score, but generative scores mitigate the problem to some extent. 

\subsection{Sensitivity to syntax and contents}
Based on the observation that ViusalGPTScore mitigates the bags-of-words problem to some extent, we are curious about whether they lean more towards evaluating syntactic correctness than content relevance when assessing the compositionality of GVLMs. 
To examine it, we design an experiment using the test set of Flickr30K dataset~\cite{flickr30k}.
Specifically, we sample 507 image-text pairs and construct three types of evaluation cases as shown in Fig.~\ref{fig:syntax}. 
Given an image, the task is to retrieve the positive reference from the cases below. The final scores are averaged over 507 test samples. 
In \texttt{Case 1}, each positive reference sentence is accompanied by two hard negatives with shuffled nouns, adjectives and trigrams.  
In \texttt{Case 2}, the provided negatives are fluent and syntactically correct captions sampled from COCO, which are unrelated to the visual contents.
In \texttt{Case 3}, we keep only adjectives and nouns in the positive reference sentences by removing all the adverbs, pronouns and modifiers. %

We present Recall@1 of VisualGPTScore for the GVLM (LLaVA-7B), and vision-language similarity for the EVLM (CLIP) in three evaluation cases. 
As shown in Fig.~\ref{fig:syntax_chart}, 
the LLaVA model can easily discriminate the right reference sentences from the shuffled ones, reaching 98.62\% with the help of VisualGPTScore. 
However, if the negatives are random reference sentences in \texttt{Case 2},
the performance degradation is up to 31.56\%. 
In \texttt{Case 3}, where the sentences are syntactically incorrect, the performance drops to 27.02\%.
In contrast, CLIP excels at excluding negative sentences that are contextually unrelated to the image, but suffers from insensitive to syntax and sentence order.   

The potential reason for the above results is the difference in the pre-training paradigm.
Specifically, the generative model pre-training is to maximize the likelihood of the next token prediction in an auto-regressive manner.
In contrast, the training objective of EVLM is to maximize the alignment between positive image-text pairs and minimize that between negative ones. 
Previous research~\cite{whenandwhy} shows that CLIP takes the short-cut strategy of not encoding the order information, but only object features for retrieval/captioning tasks, which conforms to our finding. 
We also believe that the generative VLMs take the short-cut strategy of not fully mapping the visual and linguistic features, but leveraging the emerging capacity of LLM part to generate based on limited visual cues. \textit{This reliance on the LLM part results in a bias towards syntactical correctness in captions under the criteria of generative score.}

\section{Benchmarks Examination}
\label{sec:dataset}

\begin{figure*}[]
\centering
\includegraphics[width=1.0\linewidth,trim={0cm 0cm 0cm 0cm}]{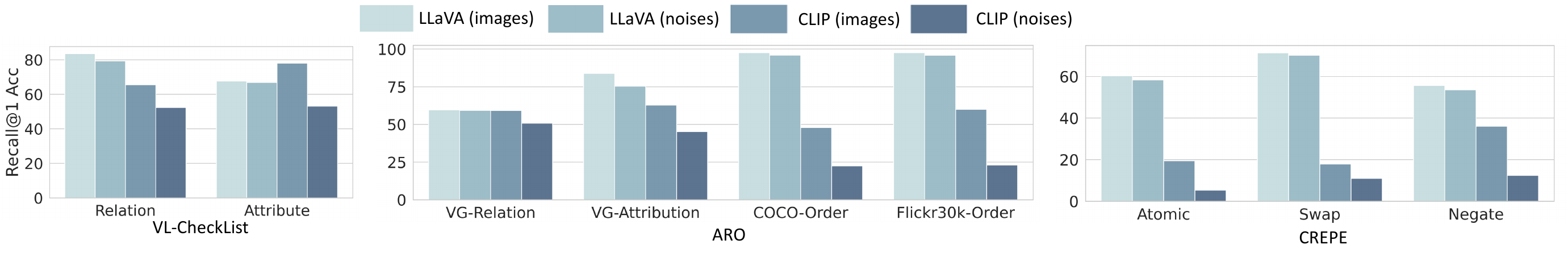}
\caption{The drop in performance of the LLaVA model when performing compositional reasoning on \textbf{nonsensical noisy images} is minimal in existing benchmarks, whereas the CLIP model exhibits a significant decrease. This indicates current benchmarks are exploited by the LLM part of GVLMs, not effective in measuring the multimodal compositionality.} 
\label{fig:noise}
\end{figure*}

From above, we know current benchmarks are curated for evaluating EVLMs based on similarity score originally. Hence, we examine the impact of using these datasets for evaluating GVLMs with VisualGPTScore, and uncover the bias of existing datasets. 

\subsection{Syntactical bias in current benchmarks}
According to the observation made in Section~\ref{sec:evaluation}, it is evident that auto-regressive vision-language models exhibit sensitivity toward the syntax and order of phrases. 
Hence, existing benchmarks that generate hard negatives by swapping, shuffling, or replacing specific entities promote a syntactical bias, which refers to a preference for models to rely on the morphological structure of words. 
Consequently, this bias can be exploited by GVLMs to effortlessly differentiate between positive and negative samples.

To show that the bias exists in current compositional reasoning benchmarks, we conduct the ablation of utilizing both GVLMs and EVLMs to reason nonsensical images with normal reference sentences. 
Specifically, we construct the image-text pairs by replacing the original images with images composed of \textit{random noises}.
We observe the performance drop in both the GVLMs and EVLMs. 
As shown in Fig.~\ref{fig:noise}, the performance degradation of CLIP (ViT-B/32) is large, approaching the \texttt{Recall@1} accuracy of randomly choosing. 
However, as for the LLaVA-7B, the trend of performance dropping is not obvious, indicating the GVLMs make the right choices \textbf{solely} based on the linguistic reference sentences without visual features. Therefore, almost all the benchmarks lean towards evaluating the linguistic part of GVLMs, rather than the visio-linguistic understanding of GVLMs.

\subsection{SyntaxBias Score}
\begin{figure}[!h]
\centering
\includegraphics[width=1.0\linewidth,trim={0cm 0cm 0cm 0cm}]{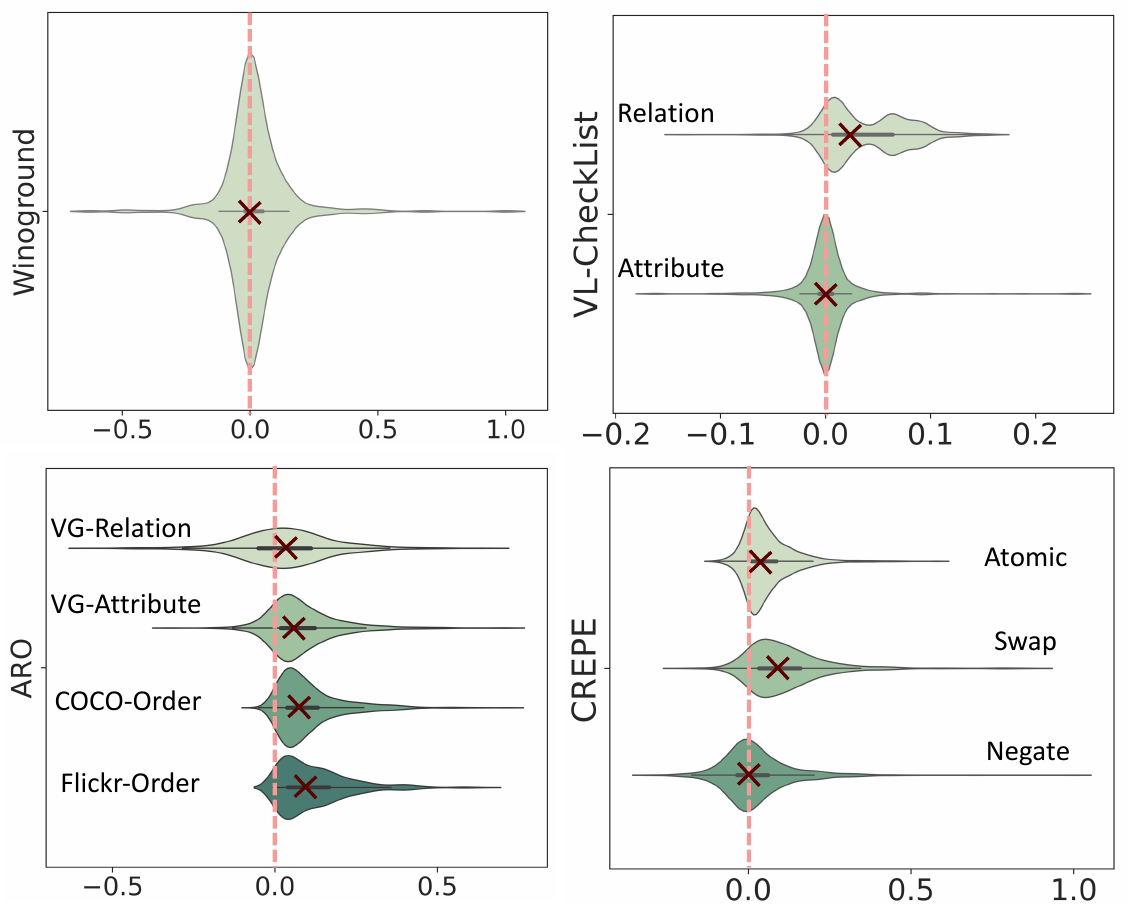}
\caption{We visualize the distribution of \textit{SyntaxBias Score} in current benchmarks. The \textit{SyntaxBias Score} is defined as the difference between the LLM-based generative scores of positive and negative references. 
For ARO, VL-CheckList and CREPE, the distribution of the SyntaxBias Scores is situated towards the positive end (to the right of the red line), implying that these benchmarks are biased to positive captions syntactically.}
\label{fig:bias}
\end{figure}

To alleviate the syntactical bias in current benchmarks, we first quantify the bias for analysis. 
In an ideal scenario, in the absence of visual intervention, the quantified scores generated by GVLMs for positive and negative reference sentences should be equivalent. 
Therefore, we define the SyntaxBias Score to measure the syntactical discrepancy between positive and negative reference sentences.
Formally, the SyntaxBias Score is calculated using the generative scores of positive and negative text produced by auto-regressive language models:
\begin{align}
    \label{eqn:syntaxbias}
    & Score_{SyntaxBias} \nonumber \\ 
    & = \Delta(\sum\limits_{i=1}\limits^{m}w_i {\rm log}P(p_i|\bm{p}_{<i}; \theta)  \\
    & - \sum\limits_{j=1}\limits^{n} \hat{w}_j {\rm log}P(n_j|\bm{n}_{<j}; \theta)),  \nonumber
\end{align}
where $\Delta, \mathbf{p}, \mathbf{n}, \theta$ represent normalization, positive tokens, negative tokens, and parameters of LLMs respectively.
we leverage a strong LLM, Vicuna-13B-v1.5~\cite{vicuna}, to compute the SyntaxBias Score,
which are normalized between $-1$ and $1$. 
We present the visualization of SyntaxBias Score distributions over different benchmarks in Fig.~\ref{fig:bias}.
We find that most of the mainstream benchmarks except Winoground are biased towards positive captions with distribution centers located to the right, \textit{which makes the generative scores of GVLMs on these benchmarks overvalued.}

\section{Mitigate the Bias in Benchmarks}

\begin{table*}[!h]
\centering
\scriptsize
 \resizebox{\textwidth}{!}{
\begin{tabular}{cccccccccc}
 &Comprehensive  & \multicolumn{2}{c}{Relation} &\multicolumn{2}{c}{Attribute}  &Atomic &Negate &\multicolumn{2}{c}{Content}    \\ 
  \cmidrule(r){2-2} \cmidrule(r){3-4} \cmidrule(r){5-6}  \cmidrule(r){7-7} \cmidrule(r){8-8} \cmidrule(r){9-10} 
  & Winoground &VL-CheckList &VG(ARO) &VL-CheckList &VG(ARO) &VG(CREPE) &VG(CREPE) &COCO &Flickr30K \\ 
  \midrule
 num of images &800 &5,193 &2,328 &5,858 &5,193 &1,954 &1,930 &2500 &500\\
num of references &800 &10,386 &4,656 &11,716 &10,386 &11,724 &11,580 &7,500 &1,500\\ \cmidrule{3-10}
metrics &Group Score &\multicolumn{8}{c}{Recall@1} \\ \cmidrule{3-10}
random results &16.7\% &50.0\% &50.0\% &50.0\% &50.0\% &16.7\% &16.7\% &33.3\% &33.3\% \\ \midrule
\multicolumn{10}{c}{\textit{Human Evaluation (closer to 0 is better)}} \\
origin ref. &- &3.18 &1.73 &0.95 &3.29 &1.67 &2.11 &- &- \\
SADE ref. &- &1.40 &0.62 &0.35 &1.01 &0.94 &1.63 &- &-\\
\end{tabular}
}
\caption{Taxonomy of \textbf{SADE} benchmark and human evaluation results on rating bias. Each branch undergoes human evaluation based on 50 reference sentences from the original dataset and 50 from SADE. }
\label{tab:data}
\end{table*}

In this section, we propose a strategy to modify the benchmarks and mitigate the syntactical bias to provide a better evaluation of GVLMs. %
Specifically, we filter current datasets leveraging LLMs and add a novel challenge to evaluate visual content understanding. 
We name the new benchmark as \textbf{S}ynt\textbf{A}ctical \textbf{De}-biased benchmark, abbreviated as \textbf{SADE}.
In the following, we describe the filtering details of each dataset and the new challenge. Then we show human evaluation to show the effectiveness of SADE.

\subsection{Winoground}
The Winoground~\cite{winoground} dataset comprises 400 image-text pairs, with each pair consisting of two images and two captions. 
The two captions exhibit identical sets of morphemes, albeit in different orders. 
Different from other benchmarks that construct hard negatives by simply altering the positive texts, both positive and negative texts in Winoground are fluent, meaningful, and can match related images. 
Thus, we include all samples in Winoground into the SADE benchmark without further mitigation, aiming to evaluate the comprehensive multimodal compositional understanding of GVLMs, especially on the \textit{pragmatics, symbolic and series} factors as introduced in~\cite{winoground}.

\begin{table*}[htbp]
\centering
\small
\renewcommand\arraystretch{1.2}
\begin{tabular}{cccccccccc}
\toprule
 &Comprehensive  & Relation &Attribute  &Atomic &Negate &Content    \\ 
  \midrule
 LLaVA-7B~\cite{llava} &13.00 &65.52 &70.55 &35.01 &59.01  &42.02 \\
 LLaVA-13B~\cite{llava} &17.00 &62.75 & 72.70 &38.33 &7.56 &49.80\\
 MiniGPT-7B~\cite{minigpt4} &9.50 &66.18 &78.48 &35.62 &24.15 &19.92\\
 mPLUG-Owl~\cite{mplug} &11.00 &65.91 &69.04 &34.90 &54.61 &35.73\\
 InstructBLIP~\cite{instructblip} &\textbf{26.00} &\textbf{73.87} &79.39 &44.37 &66.84 &\textbf{57.83} \\
 LLaMA Adapter V2~\cite{llama-adapter-v2} &7.75 &58.67 &65.07 &31.32 &20.26 &10.48 \\
 Emu~\cite{emu} &4.00 &68.54 &\textbf{85.84} &\textbf{51.38} &\textbf{87.20} &2.79\\
\bottomrule
\end{tabular}
\caption{Evaluation results of GVLMs on SADE benchmark. All the models are instruction-tuned. We present the average performance of two sub-branches within the categories of \textit{Relation}, \textit{Attribute} and \textit{Content}.}
\label{tab:more_results}
\end{table*}

\subsection{Relations and attributes}
Real-world natural scenes are inherently intricate, encompassing a multitude of specific attributes such as colors, materials, and object relationships. 
Models that can tackle compositional reasoning require a nuanced understanding that goes beyond mere object-level analysis. 
Hence, we collect relation and attribute branches from ARO~\cite{ARO} and VL-CheckList~\cite{vl-checklist}. 
To mitigate the syntactical bias, we compute the SyntaxBias Score of the samples as described in Eqn. ~\ref{eqn:syntaxbias} and filter out ones that have a higher score than the threshold. 
The idea is to ensure that samples with strong syntactical bias are excluded for better vision-language compositional evaluation. 

We choose the filtering thresholds of the SyntaxBias Score to be close to zero (specifically, by ensuring the $p-value$ of the SyntaxBias Score is statistically below $1e-5$).
The filtered data includes 5,193 items from VL-CheckList and 2,328 items from Visual Genome~\cite{vg} to measure relation reasoning, and 5,858 items from VL-CheckList as well as 5,193 items from Visual Genome to evaluate attribute reasoning. 
Specifically, for VL-CheckList, the \textit{Relation} branch contains two subclasses, \textit{i.e.} \textit{action} and \textit{spatial}, and the \textit{Attribute} branch includes \textit{action}, \textit{color}, \textit{material}, \textit{size} and \textit{state}. 
The number of items in each subclass is elaborated in Table~\ref{tab:data}.

\subsection{Atomic and negate}
In CREPE benchmark~\cite{crepe}, the authors propose to assess the VLMs on captions that atoms are replaced or negated. 
The atom replacing is like \sethlcolor{lightgray}\hl{\textit{a \textbf{bus} with a side, light, and window}} \textit{v.s.} \sethlcolor{lightgray}\hl{\textit{a \textbf{train} with a side, light, and window}}, whereas the atom or sentence negating is as \sethlcolor{lightgray}\hl{\textit{Another bowl on a cloth with an orange in it. 
The another bowl has a reflection and \textbf{casts a shadow}}} \textit{v.s.} \sethlcolor{lightgray}\hl {\textit{Another bowl on a cloth with an orange in it. 
The another bowl has a reflection and casts something. \textbf{There is no shadow}}}. 
There is a considerable proportion of reconstructed captions in CREPE that are fluent and coherent, thereby we also leverage the same method to filter the samples as we do for relations and attributes.

\subsection{Replace syntactic perturbation with a content-only understanding challenge}
A plethora of benchmarks perturbs the order information in the reference sentences to measure the word order sensitivity of EVLMs, which tend to treat the captions as \textit{bags of words} as we present in Fig.~\ref{fig:bag_of_words}. 
The hard negative construction methods include swapping atoms, shuffling nouns, adjectives, trigrams, and all words \textit{etc}. 
However, due to the intrinsic syntactical awareness of LLMs, the challenge of order perturbation is not effective in assessing the visio-linguistic compositionality of GVLMs. 
Hence, we abandon the order challenge and propose a content-only understanding challenge.

Specifically, we modify the positive reference sentences from COCO~\cite{coco} and Flickr30K~\cite{flickr30k}, keeping only the object- and attribute-related atoms/words. 
Then, we randomly select fluent, coherent and meaningful reference sentences from other datasets to serve as hard negatives, which are unrelated to the visual content. 
Examples of this challenging task can be found in Fig.~\ref{fig:supple} in the Appendix.
\textit{The task poses a challenge and exemplifies the robustness of GVLMs against their inherent inclination towards syntactically correct reference sentences.}

\subsection{Human evaluation of SADE}
In order to illustrate that our proposed SADE alleviates the syntactical bias, we ask two annotators to rate the disparity between positive and negative reference sentences. 
The rating score ranges from -5 to 5, where the higher the score, the more reasonable text is for positive reference sentences.
Conversely, the lower the score, the more reasonable the text is for negative ones. 
The definition of \textit{reasonable} comprises fluency, syntax, and the meaning of sentences. 
Note the reference sentences from the original dataset or SADE are agnostic to the annotators and we average the ratings of them. 
Table~\ref{tab:data} clearly demonstrates that the reference sentences in our SADE benchmark substantially mitigate bias, as indicated by the score of human judgments approaching zero. 
The drop implies that the syntactical disparity between positive and negative reference sentences is drastically narrowed.

\subsection{Results of GVLMs on SADE}
Based on the SADE benchmark, we report the performance of more concurrent GVLMs based on the VisualGPTScore metric in Table~\ref{tab:more_results}.
It can be observed that InstructBLIP~\cite{instructblip} and Emu~\cite{emu} hold the top-2 positions in almost all dimensions of our benchmark. 
However, the abysmal performance on \textit{Comprehensive} and \textit{Content} implies the vulnerability of Emu when negative reference sentences are hard and challenging. 
In contrast, InstructBLIP and LLaVA-13B~\cite{llava} are more robust to the \textit{Content} challenge and achieve high performance on hard negatives.
This provides the first de-biased and comprehensive evaluation of recent GVLMs in terms of visual compositionality. 
\textbf{Note that we do not claim that SADE can better measure the performance of GVLMs in all aspects.} However, it can better measure their compositionality with less syntactical bias, which is supported by the reduction of SyntaxBias Score and the human evaluation in Table~\ref{tab:data}. 
We believe this benchmark can facilitate a unified and fair comparison for future GVLM research.

\section{Conclusion}
In this work, we evaluate the compositionality of ``bridge-architecture" generative VLMs via generative multimodal score, VisualGPTScore. 
We examine both the VisualGPTScore and current benchmarks for evaluating the multimodal compositional understanding of GVLMs.
Based on the examinations, we identify the syntactical bias that exists in current datasets for GVLMs, and define the bias with SyntaxBias Score quantitatively. 
We then propose a SADE benchmark that mitigates the syntactical bias and provides a better content understanding evaluation for GVLMs.
We report the results of multiple GVLMs on our proposed SADE benchmark and uncover new findings of the GVLMs' capabilities.

\section{Limitations}
We discuss the potential limitations of this paper from two aspects. First, our proposed novel benchmark cannot be proved to better measure the performance of generative VLMs in all aspects, including emergent capability, vision understanding and complex reasoning. Our benchmark just evaluates the GVLMs in terms of VL compositionality more fairly by removing the syntactical bias in previous benchmarks.
Second, our new benchmark is based on filtering the previous ones, and sampling from them to lower the SyntaxBias Score. Thus, the scale of the whole dataset is relatively small, limiting the generalization of the benchmark to some extent.

\section{Acknowledgements}

This work was supported by the National Natural Science Foundation of China (No. 62306257) and the Guangzhou Municipal Science and Technology Project (No. 2024A04J4390).
This work was also supported by the Meituan Academy of Robotics Shenzhen.
The views and conclusions contained herein are those of the authors and should not be interpreted as necessarily representing the official policies or endorsements, either expressed or implied, of the National Natural Science Foundation, Meituan, or the Guangzhou Government.

\bibliography{custom}

\appendix

\section{Appendix}
\label{sec:appendix}

\subsection{Formulations of GVLMs and EVLMs}
\label{subsec:formulation}
In accordance with the discussion in the main text, we define GVLMs as models that combine visual encoders with large language models (LLMs) trained on large text corpora.
The visual tokens are mapped into a lexical embedding space and harnessed to generate textual content in an autoregressive manner.
Formally, given an image $I$ and the visual encoding $g(I)$ from encoders like Vision Transformer~\cite{vit}, the mapping process can be formulated as:
\begin{equation}
   \bm{z} = \mathbf{M} (g(I)), \bm{z}=\{z_1, z_2, ... , z_N\},
\end{equation}
where $N$ is the number of visual tokens and $\mathbf{M}$ is the mapping layers.
Different from EVLMs, the training objective of multi-modal auto-regressive training is to maximize the log-likelihood of the next true token. 
Denote the tokenized instructions as $\bm{p}$ and the output words as $t_i, (1\leq i \leq K)$, the GVLM training objective is defined as:
\begin{equation}
    \mathop{{\rm max}}\limits_{\theta_M, \theta_{\sigma}} \sum\limits_{i=1}^{K} {\rm log} P(t_i|\bm{p}, \bm{z}, t_1, t_2, ... ,t_{i-1}; \theta_M, \theta_{\sigma})
\end{equation}
where $\theta_M$ refers to the learnable parameters of mapping layers $\mathbf{M}$ and $\theta_{\sigma}$ refers to other tunable parameters like adapter layers in LLaMA-Adapter V2~\cite{llama-adapter-v2}, or visual abstractor and LoRA in mPLUG-Owl~\cite{mplug}.

In comparison, the training objective of EVLMs is based on the ITC or ITM loss between vision and language. Given an input image $I$ and text $T$, the encoded visual and linguistic features are denoted as $f_v$ and $f_t$. Then, two transformation matrices $W_v$ and $W_t$ are employed to project the visual and text features into a joint feature embedding space, which is formulated as:
\begin{equation}
    v = \frac{W_v^\top f_v}{||W_v^\top f_v||}, u=\frac{W_t^\top f_t}{||W_t^\top f_t||}
\end{equation}
In the shared embedding space, ITC loss narrows the discrepancy of vision and language, aligning the image-text pairs in the same batch. The training objective of this process comprises two components, \textit{i.e.} $\mathcal{L}_{v \rightarrow t}$ for text retrieval and $\mathcal{L}_{t \rightarrow v}$ for image retrieval. The similarity of matched pairs will be maximized while unmatched ones will be minimized. The formula is:
\begin{equation}
    \begin{aligned}
    &\mathcal{L}_{ITC} = \mathcal{L}_{v \rightarrow t} + \mathcal{L}_{t \rightarrow v} \\
    =& -\frac{1}{|\Omega_{v}^{+}|} \sum\limits_{T_j\in \Omega_{v}^{+}} {\rm log}\frac{{\rm exp}(v_i^{\top} u_j/\tau)}{\sum_{T_k\in \Omega_t}{\rm exp}(v_i^\top u_k/\tau)} \\
    &-\frac{1}{|\Omega_{t}^{+}|} \sum\limits_{I_i\in \Omega_{t}^{+}} {\rm log}\frac{{\rm exp}(u_i^{\top} v_j/\tau)}{\sum_{I_k\in \Omega_v}{\rm exp}(u_i^\top v_k/\tau)}
\end{aligned}
\end{equation}
where $\Omega_{v}, \Omega_{t}$ represent a batch of images and texts while $\Omega_{v}^{+}, \Omega_{t}^{+}$ denote positive subsets matched to image $I_i$ and text $T_i$.
ITM loss is a binary classification loss based on the joint representation of visual and linguistic features. Compared with ITC loss, ITM loss does not maximize the distance between negative pairs.

\begin{figure}[h!]
\centering
\includegraphics[width=1.0\linewidth,trim={0cm 0cm 0cm 0cm}]{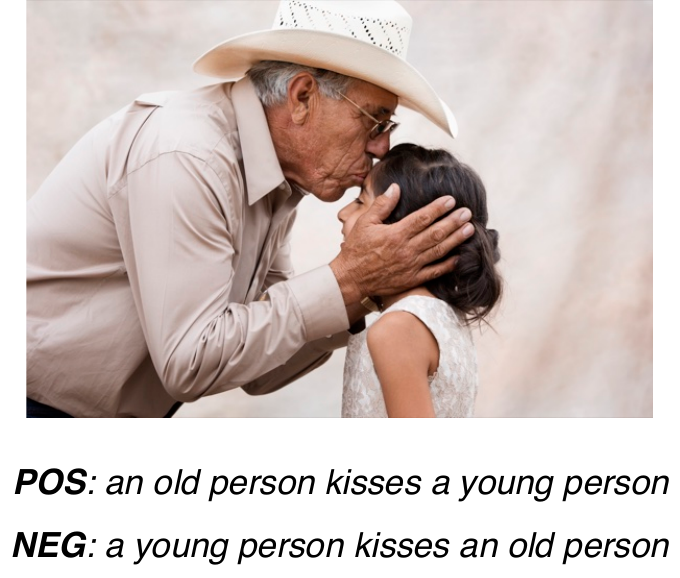}
\caption{An data example in current benchmarks. The image, positive and negative references are from Winoground~\cite{winoground}.}
\label{fig:example}
\end{figure}

\subsection{Granularity influence of VisualGPTScore.}
\begin{figure}[h!]
\centering
\includegraphics[width=0.8\linewidth,trim={0cm 0cm 0cm 0cm}]{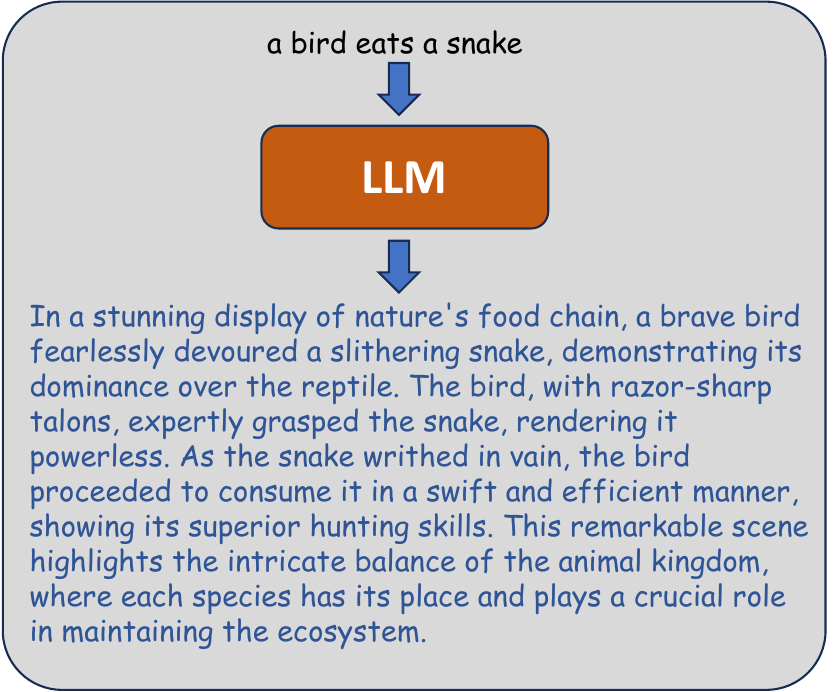}
\caption{An LLM is leveraged to fine-grain the references.}
\label{fig:granularity}
\end{figure}

To explore the influence of granularity of references in the visio-linguistic compositional reasoning, we leverage a language model to enrich the object details and relational phrases for short references in Winoground dataset, where all references are fluent and reasonable. 
Vicuna-13B-v1.5\footnote{https://huggingface.co/lmsys/vicuna-13b-v1.5/tree/main} is adopted as the LLM, which is instruction-following tuned based on LLaMA 2~\cite{llama2}, one of the strongest open-source LLMs currently. Note that we artificially filter out nonsensical and unrelated expanded captions that are not relevant to the image and keep 282 of 400 image-text pairs finally. 
The expandation of references is shown in Fig.~\ref{fig:granularity}.

\begin{table}[]
\centering
\small
 \resizebox{\columnwidth}{!}{
\begin{tabular}{cccc}
\toprule
 Models\&References &Text Score &Image Score &Group Score \\
 \midrule
LLaVA+Original &12.06 &12.77 &7.45 \\
LLaVA+Fine-grained &8.51({\color{green}-3.55}) &37.23({\color{red}+24.50}) &6.38({\color{green}-1.07}) \\
MiniGPT-4+Original &18.44 &17.02 &9.22 \\
MiniGPT-4+Fine-grained &6.03({\color{green}-12.41}) &31.91({\color{red}+14.89}) &4.96({\color{green}-4.26}) \\
\bottomrule
\end{tabular}
}
\caption{Accuracy of LLaVA and MiniGPT-4 on original and fine-grained references of filtered Winoground dataset. The definitions of Text Score, Image Score, and Group Score is specified in Winoground~\cite{winoground}.}
\label{tab:granularity}
\end{table}

We present the results in Table~\ref{tab:granularity}, and observe that the performance of ``Image Score" has been largely improved, indicating the fine-grained references are beneficial for text-to-image retrieval based on the definition of ``Image Score" in Winoground.

\subsection{Zero-shot answer generation}
Unlike EVLMs, GVLMs excel in zero-shot generation when guided by instructions, prompts, or demonstrations. 
We attempt to prompt and demonstrate the LLaVA and MiniGPT-4 to output the choices of positive or negative reference sentences based on corresponding images. 
However, we do not consider zero-shot generation of answers in our paper with two reasons. First, zero-shot answer generation cannot reflect the GVLMs' compositional understanding quantitatively, without scores or probabilities to show the confidence of judgements. 

Second, when demonstrating the GVLMs to generate the option number of reference sentences directly, it is hard to acquire the direct answer due to the free-form answer format, especially considering the emergent capability is limited in relatively small-scaled GVLMs. 
In a limited number of instances, we observed successful model outputs where options or inference processes were accurately provided, resembling the blue line in Table~\ref{tab:free-form}. However, in the majority of cases, the GVLMs generated fabricated answers that were characterized by a rhetorical tone, similar to the examples shown in Table~\ref{tab:free-form}. Also, there are cases that the rationales are correct, but the option number is wrong, conflicting with the reasoning process of GVLMs (shown in orange line in Table~\ref{tab:free-form}).
Hence, assessing the compositionality of GVLMs solely through direct zero-shot answer generation becomes challenging, particularly when the zero-shot capability is constrained within a relatively small-scale model like the 7B variant. Furthermore, it is not possible to quantitatively analyze the alignment of a single image-text pair using this type of evaluation method. 

\begin{table}[h!]\centering

\begin{minipage}{1.0\columnwidth}\vspace{0mm}    \centering
\begin{tcolorbox} 
    \centering
    \small
     \hspace{-6mm}
\begin{itemize}[leftmargin=7.5mm]
\setlength{\itemsep}{2pt}
\item \textcolor{blue}{Ducks are known to be efficient at swimming, and they spend a significant amount of time in the water. In this image, a duck is swimming on the water while facing the camera. The option B is most similar to the image since it describes a duck walking or swimming in the water.}
\item \textcolor{teal}{Which of the two options provides a better visual representation of the actual image?}
\item \textcolor{teal}{Which character should I output?}
\item \textcolor{teal}{Which one is most similar to the image?}
\item \textcolor{orange}{The girl is standing to the left of the giraffe, and they seem to be in a natural environment. The girl is feeding the giraffe from her hand, and the giraffe has its tongue out, reaching towards her hand. The answer is more likely to be B.}
\end{itemize}

\end{tcolorbox}
    
\vspace{-2mm}
\caption{Examples of zero-shot answer generation method. \textcolor{blue}{Blue:} free-form generation, \textcolor{teal}{Teal:} fabricated answers, \textcolor{orange}{Orange:} conflicting rationales and answers. }
    \label{tab:free-form}
\end{minipage}
\end{table}

\subsection{Examples of content challenge of SADE}
\begin{figure*}[htbp]
\centering
\vspace{-10pt}
\includegraphics[width=1.0\linewidth,trim={0cm 0cm 0cm 0cm}]{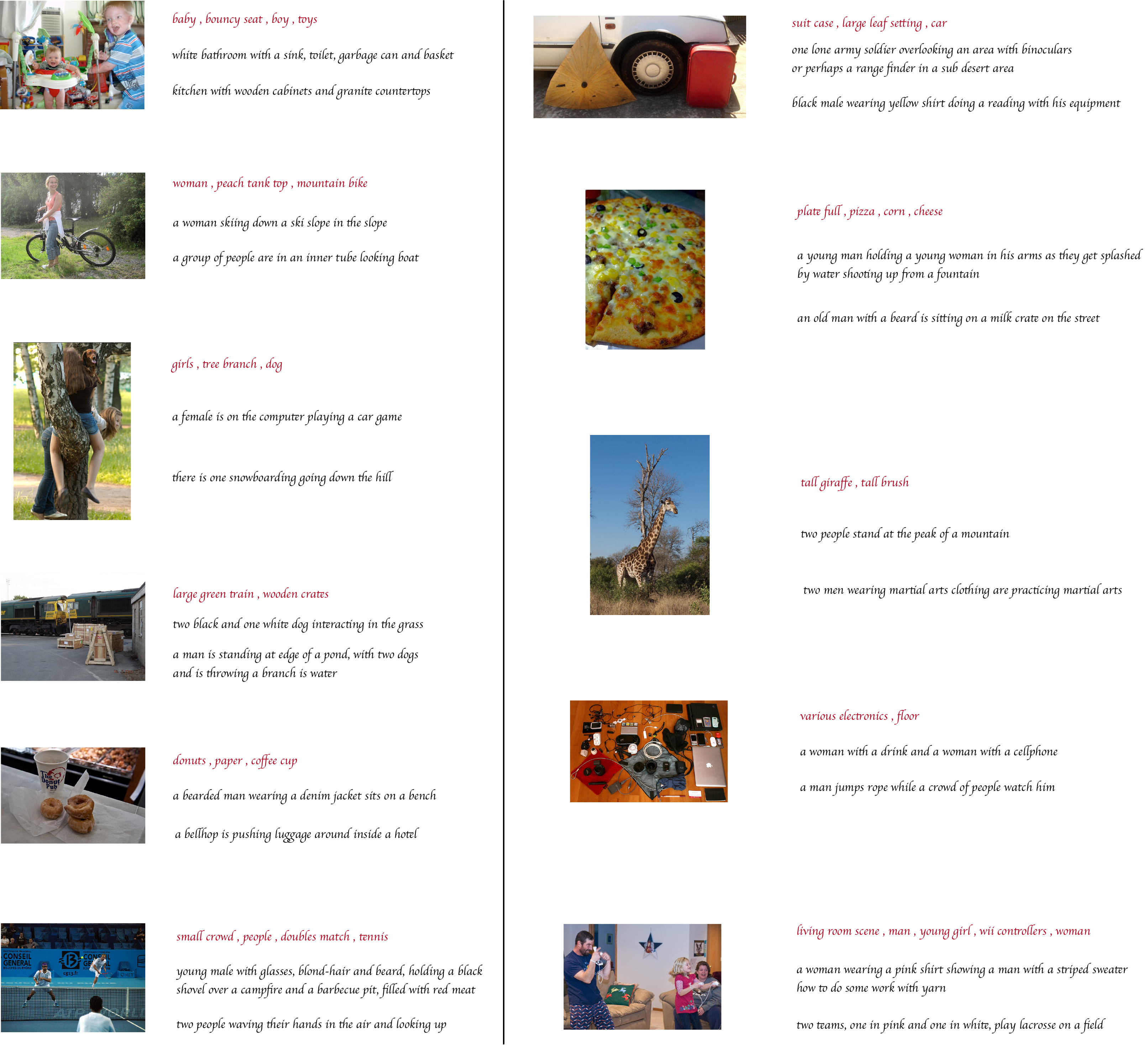}
\vspace{-10pt}
\caption{Examples of Content challenge in our SADE benchmark. The red texts denote positive reference sentences that solely capture visual elements while disregarding sentence structure. On the other hand, the black texts represent negative reference sentences that are grammatically sound and meaningful, yet unrelated to the visual contents depicted in the images. 
}

\label{fig:supple}
\end{figure*}

We present some examples of items in the Content challenge branch in our SADE benchmark in Fig.~\ref{fig:supple}. Each item comprises one positive reference sentence and two negative ones. The red texts are positive reference sentences that only kept visual content-related phrases, while the black texts are negative reference sentences that were extracted randomly from other datasets. The negative reference sentences are fluent, coherent and meaningful, but irrelevant to the contents of the images.

The pure content understanding is challenging. Specifically, the intrinsic inclination of GVLMs towards syntactic correctness drives the GVLMs to prefer negative reference sentences. From the perspective of our proposed SyntaxBias Score, the bias of our Content Challenge is opposite to the current benchmarks, which is biased to the negative reference sentences in syntax. Therefore, GVLMs have to overcome the negative bias in syntax and show the robustness of visual understanding.

\end{document}